\documentclass[11pt,a4paper]{article}
\usepackage[hyperref]{acl2021}
\usepackage{times}
\usepackage{latexsym}

\usepackage{microtype}

\usepackage{graphicx}
\usepackage{booktabs}
\usepackage{multicol}
\usepackage{multirow}
\usepackage{amsmath}
\usepackage{amssymb}
\usepackage{appendix}

\aclfinalcopy 
\def\aclpaperid{3689} 

\title{
Incorporating Connections Beyond Knowledge Embeddings: \\ 
A Plug-and-Play Module to Enhance Commonsense Reasoning \\ 
in Machine Reading Comprehension
}

\author{
Damai Dai\textsuperscript{1},
Hua Zheng\textsuperscript{1},
Zhifang Sui\textsuperscript{1,2},
Baobao Chang\textsuperscript{1,2},
\\
\textsuperscript{1}Key Lab of Computational Linguistics, School of EECS, Peking University 
\\
\textsuperscript{2}Peng Cheng Laboratory, China 
\\
\texttt{\{daidamai,zhenghua,szf,chbb\}@pku.edu.cn}
}

\date{}

\begin{document}
\maketitle

\begin{abstract}

Conventional Machine Reading Comprehension (MRC) has been well-addressed by pattern matching, but the ability of commonsense reasoning remains a gap between humans and machines. 
Previous methods tackle this problem by enriching word representations via pre-trained Knowledge Graph Embeddings (KGE). 
However, they make limited use of a large number of connections between nodes in Knowledge Graphs (KG), which could be pivotal cues to build the commonsense reasoning chains. 
In this paper, we propose a \textbf{P}lug-and-play module to \textbf{I}ncorporat\textbf{E} \textbf{C}onnection information for commons\textbf{E}nse \textbf{R}easoning (\textbf{PIECER}). 
Beyond enriching word representations with knowledge embeddings, PIECER constructs a joint query-passage graph to explicitly guide commonsense reasoning by the knowledge-oriented connections between words. 
Further, PIECER has high generalizability since it can be plugged into suitable positions in any MRC model. 
Experimental results on ReCoRD, a large-scale public MRC dataset requiring commonsense reasoning, show that PIECER introduces stable performance improvements for four representative base MRC models, especially in low-resource settings.\footnote{We will release our code once this paper is officially accepted. }

\end{abstract}

\section{Introduction}

Machine Reading Comprehension (MRC) is a pivotal and challenging task in Natural Language Processing (NLP), which aims to automatically comprehend a context passage and answer related queries. 
In recent years, especially after the proposal of various large-scale Pre-Trained Models (PTM)~\citep{bert,roberta}, MRC has achieved great success on multiple datasets such as SQuAD~\citep{squad} and NewsQA~\citep{newsqa}. 
However, current state-of-the-art models perform significantly worse on ReCoRD~\citep{record}, a large-scale MRC dataset that is more challenging since it requires commonsense reasoning. 
This suggests that current excellent performances of MRC models may be due to the strong ability of pattern matching~\citep{pattern_matching1,pattern_matching2,pattern_matching3}, but they still lag behind in the ability of commonsense reasoning. 

\begin{figure}[t]
\centering
\includegraphics[width=0.99\linewidth]{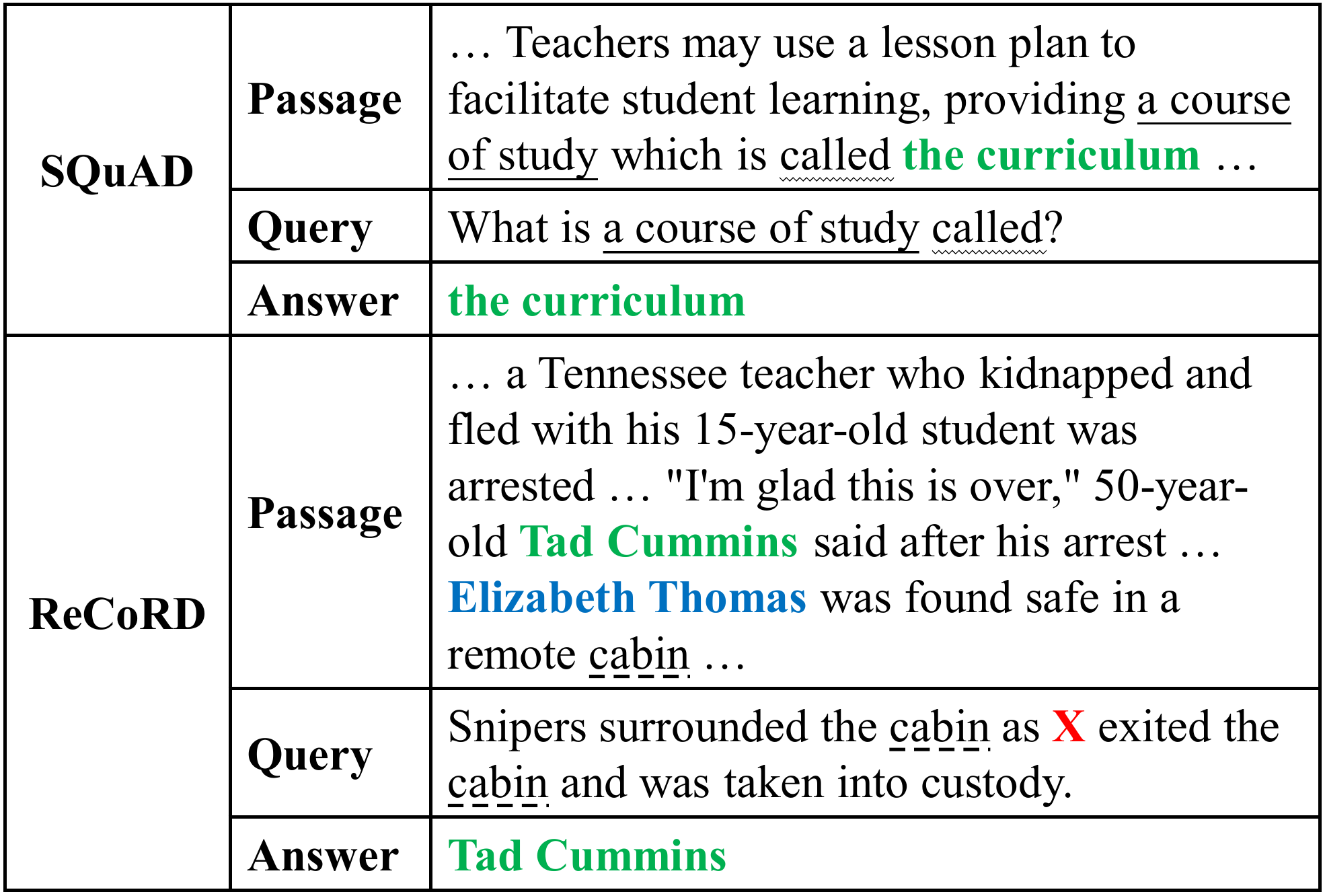}
\caption{
Two examples selected from SQuAD and ReCoRD. 
The SQuAD example requires only simple pattern matching, while the ReCoRD example requires commonsense reasoning. 
We mark the same keywords using the same underlines.
}
\label{fig:example}
\end{figure}

Figure~\ref{fig:example} illustrates the difference between SQuAD and ReCoRD. 
In the SQuAD example, both keywords in the query (``a course of study'' and ``called'') directly appear around the correct answer in the passage, so we can easily find the answer \textbf{\textit{the curriculum}} by pattern matching. 
By contrast, the ReCoRD example, where we need to select a candidate entity (marked in bold) in the passage to replace the placeholder (denoted by a bold red \textbf{X}) in the query, requires commonsense reasoning. 
Firstly, few keywords (only ``cabin'') in the query are covered by the passage. 
Secondly, there are confusing entities (\textbf{\textit{Elizabeth Thomas}} as the hostage vs. \textbf{\textit{Tad Cummins}} as the kidnapper) that require external knowledge for differentiation. 
Therefore, it is almost impossible to find the correct answer by only pattern matching. 
Instead, with the commonsense knowledge that ``be taken into custody'' is similar to ``be arrested'', we can infer that \textbf{\textit{Tad Cummins}} is the answer. 
These two examples reveal that pattern matching can only tackle MRC at a shallow level, while difficult datasets like ReCoRD require commonsense reasoning in depth. 

To make up for the deficiency in commonsense reasoning, previous methods attempt to leverage knowledge stored in Knowledge Graphs (KG) such as WordNet~\citep{wordnet}, NELL~\citep{nell}, and ConceptNet~\citep{conceptnet}. 
\citet{knreader} encode knowledge as a key-value memory and enrich each word by memory querying. 
\citet{kblstm} and~\citet{ktnet} enrich each word by applying attention mechanism to its KG neighbors and a sentinel vector. 
\citet{skg} propose to update the representation of each word by aggregating knowledge embeddings of its KG neighbors via graph attention~\citep{graph_attention}. 
These methods enrich each word separately by fusing pre-trained knowledge embeddings. 
However, how to explicitly leverage knowledge-oriented connections between words in KGs, which could be pivotal cues to build the commonsense reasoning chains, are barely investigated. 

In this paper, we propose a \textbf{P}lug-and-play module to \textbf{I}ncorporat\textbf{E} \textbf{C}onnection information for commons\textbf{E}nse \textbf{R}easoning (\textbf{PIECER}). 
Beyond leveraging only knowledge embeddings, PIECER constructs a joint query-passage graph to explicitly guide commonsense reasoning by the knowledge-oriented connections. 
Further, PIECER has high generalizability since it can be plugged into suitable positions in any MRC model. 
PIECER is composed of three submodules. 
The \textbf{knowledge embedding injection submodule} aims to enrich words with background knowledge by fusing knowledge embeddings pre-trained on an external KG. 
The \textbf{knowledge-guided reasoning submodule} aims to leverage knowledge-oriented connections to guide interactions between words, thus facilitating commonsense reasoning chain building. 
The \textbf{self-matching submodule} is optional, aiming to further adapt the knowledge-enhanced information to a specific MRC task. 

Our contributions are summarized as follows: 
\begin{itemize}
    \item
    Beyond leveraging only knowledge embeddings, we propose to incorporate the connection information in KGs to guide commonsense reasoning in MRC. 
    \item
    We design a plug-and-play module called PIECER for high generalizability, which can be plugged into suitable positions in any MRC model to enhance its ability of commonsense reasoning. 
    \item 
    We evaluate PIECER on ReCoRD, a large-scale public MRC dataset requiring commonsense reasoning. 
    Experimental results and elaborate analysis validate the effectiveness of PIECER, especially in low-resource settings. 
\end{itemize}

\section{Methodology}

In this section, we first formulate the MRC task and then describe our proposed PIECER in detail. 

\subsection{Task Formulation}
\label{sec:problem}

Let $\mathcal{P}$ denote a context passage consisting of $N$ words $\mathcal{P}=\left\{w^{(p)}_i\right\}^{N}_{i=1}$ and $\mathcal{Q}$ denote a query consisting of $M$ words $\mathcal{Q}=\left\{w^{(q)}_i\right\}^{M}_{i=1}$. 
Given $\mathcal{P}$ and $\mathcal{Q}$, MRC requires reading and comprehending them and then predicting an answer to the query. 
Specifically, in this paper, we tackle the extractive MRC that requires extracting a continuous span in $\mathcal{P}$ as the answer, i.e., $\text{answer} = w^{(p)}_{s:t}$, where $s$ and $t$ are the answer boundaries. 

\begin{figure*}[t]
\centering
\includegraphics[width=0.99\linewidth]{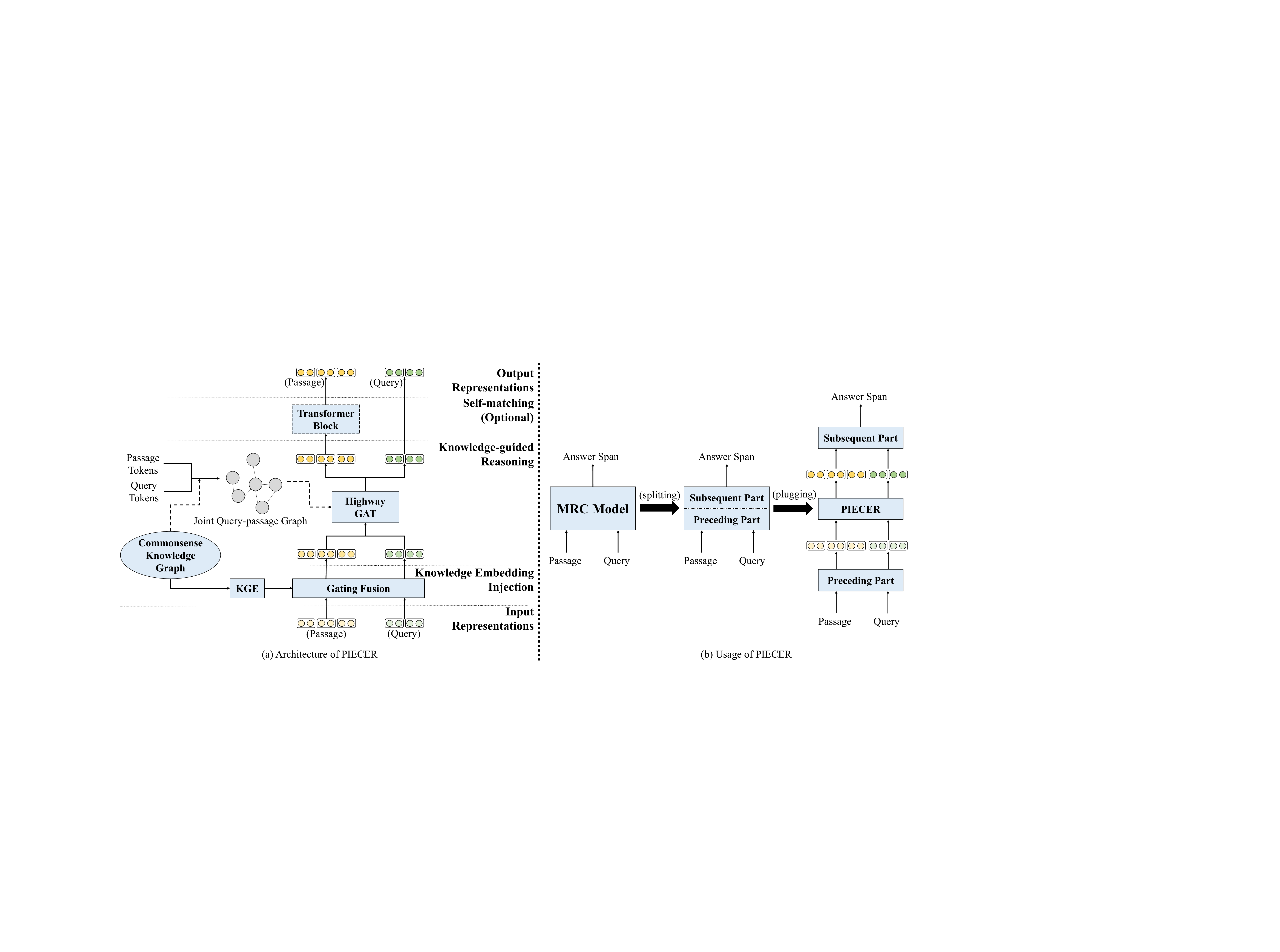}
\caption{
An overview of PIECER. 
Subplot (a) illustrates how PIECER leverages external commonsense knowledge via its submodules. 
Subplot (b) illustrates how to plug PIECER into an existing MRC model. 
}
\label{fig:model}
\end{figure*}

\subsection{Proposed Module: PIECER}

As shown in Figure~\ref{fig:model}~(a), PIECER is composed of three submodules: a knowledge embedding injection submodule, a knowledge-guided reasoning submodule, and an optional self-matching submodule. 
In this subsection, we describe them in detail. 

\subsubsection{Knowledge Embedding Injection}

In an MRC model, word representations are learned based on the data distribution in the training set and are thus limited within the dataset. 
However, commonsense reasoning usually requires background knowledge beyond the dataset as shown in Figure~\ref{fig:example}. 
To inject external background knowledge into words, this submodule adopts a gating mechanism to fuse word representations with pre-trained knowledge embeddings. 

First, we adopt Knowledge Graph Embedding (KGE) methods, such as TransE~\citep{transe} or DistMult~\citep{distmult}, to pre-train knowledge embeddings for each entity in a KG: 
\begin{equation*}
    \left\{ \mathbf{e}_i \right\} = \operatorname{KGE} (\mathcal{G}),
\end{equation*}
where $\left\{ \mathbf{e}_i \right\}$ is the entity embedding set and $\mathcal{G}$ is the selected KG. 

Then, for each word $w$ in $\mathcal{P}$ and $\mathcal{Q}$, we retrieve from $\mathcal{G}$ all unigram entities $e_i$ that have the same lemma with $w$, and adopt a gating mechanism to fuse the word representation and the retrieved knowledge embeddings: 
\begin{align*}
    \mathbf{e} &= \operatorname{Mean} (\left\{ \mathbf{e}_i|\operatorname{lemma}(e_i) = \operatorname{lemma}(w) \right\}), \\
    gate &= \sigma ( \mathbf{W}_{\mathbf{g}}[\mathbf{w};\mathbf{e}] + \mathbf{b}_{\mathbf{g}} ), \\
    \mathbf{w}^{\prime} &= \mathbf{w} \cdot gate + \mathbf{e} \cdot (1 - gate), 
\end{align*}
where $\operatorname{lemma}(\cdot)$ denotes lemmatization. 
$\mathbf{e}$ is the mean entity embedding. 
$\mathbf{w}$ is the input word representation. 
$\sigma$ denotes sigmoid. 
$\mathbf{W}_{\mathbf{g}}$ and $\mathbf{b}_{\mathbf{g}}$ are trainable parameters. 
$gate$ is the gating weight. 
$\mathbf{w}^{\prime}$ is the output word representation with background knowledge. 

\begin{figure*}[t]
\centering
\includegraphics[width=0.99\linewidth]{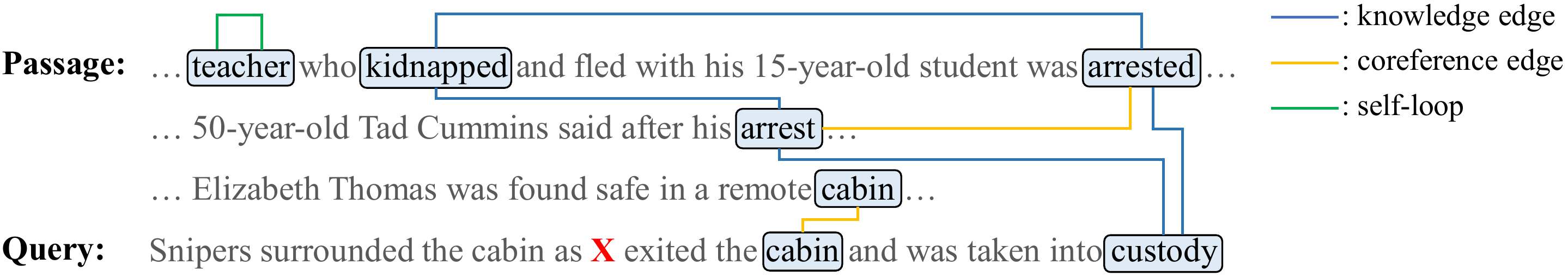}
\caption{
An illustration of a joint query-passage graph. 
For simplicity, we show only several instances for each category of edges. 
Red \textbf{X} in the query denotes the answer placeholder to be replaced by an entity in the passage. 
}
\label{fig:graph}
\end{figure*}

\subsubsection{Knowledge-guided Reasoning}

Although the knowledge embedding injection submodule injects background knowledge into each word, the connections between words are not explicitly leveraged. 
To incorporate the connection information, this submodule constructs a joint query-passage graph according to the structure of $\mathcal{G}$ and designs a Highway GAT for multi-hop reasoning. 

To construct the joint graph, we treat words in $\mathcal{P}$ and $\mathcal{Q}$ as nodes and consider three categories of edges: knowledge edge, coreference edge, and self-loop. 
For \textbf{knowledge edge}, we first link each word in the passage or query to an entity in $\mathcal{G}$ by lemma matching. 
Then, for each pair of words, if they are connected in $\mathcal{G}$, we add an edge between them. 
For \textbf{coreference edge}, we assume that two words with the same lemma are coreferential and add an edge between them. 
For \textbf{self-loop}, we add an edge between each word and itself. 
In particular, we exclude all edges connecting stop words and punctuations. 
An example of the joint query-passage graph is shown in Figure~\ref{fig:graph}. 

After constructing the joint query-passage graph, we design a \textbf{Highway GAT} that combines Highway Networks~\citep{highway} and GAT~\citep{gat} for multi-hop reasoning. 
Under the guidance of the joint graph, the Highway GAT expressly amplifies the interactions between knowledge-related nodes, which can help to build the commonsense reasoning chains. 
All nodes are updated for $L$ times. 
At the $l$-th layer, we first calculate the updated representation ${\mathbf{h}^{\prime}_i}^{(l)}$ for each node $i$ by averaging $K$ attention heads: 
\begin{align*}
    {\mathbf{h}^{\prime}_i}^{(l)} &= \frac{1}{K} \sum_{k=1}^K \sum_{j \in \mathcal{N}(i)} \alpha_{k, i, j}^{(l)} \mathbf{W}_k^{(l)} \mathbf{h}_{j}^{(l-1)}, \\
    \alpha_{k, i, j}^{(l)} &= \operatorname{Softmax}_{j}\left( e_{k, i, j}^{(l)} \right), \\
    e_{k, i, j}^{(l)} &= \sigma_\text{r} \left( {\mathbf{a}_k^{(l)}}^T \left[ \mathbf{W}_k^{(l)} \mathbf{h}_{i}^{(l-1)} ; \mathbf{W}_k^{(l)} \mathbf{h}_{j}^{(l-1)} \right] \right),
\end{align*}
where $\mathbf{h}_{i}^{(l)}$ denotes the hidden state of node $i$ at the $l$-th layer. 
Specially, we set $\mathbf{h}_{i}^{(0)}$ to the output word representation with background knowledge $\mathbf{w}_i^{\prime}$. 
$\mathcal{N}(i)$ denotes the neighbor set of node $i$ in the joint graph. 
$\operatorname{Softmax}_{j}$ denotes softmax for dimension $j$. 
$\sigma_\text{r}$ denotes LeakyReLU. 
$\mathbf{W}_k^{(l)}$ and $\mathbf{a}_k^{(l)}$ are trainable parameters. 

Then, inspired by Highway Networks, we use a highway connection to control the updating ratio to obtain the final output at the $l$-th layer $\mathbf{h}_{i}^{(l)}$: 
\begin{align*}
    \mathbf{w} &= \sigma ( \mathbf{W}_{\mathbf{h}}^{(l)}\mathbf{h}_{i}^{(l-1)} + \mathbf{b}_{\mathbf{h}}^{(l)} ), \\
    \mathbf{h}_{i}^{(l)} &= \mathbf{w} \odot \mathbf{h}_{i}^{(l-1)} + (\mathbf{1} - \mathbf{w}) \odot {\mathbf{h}^{\prime}_i}^{(l)}, 
\end{align*}
where $\mathbf{w}$ is the weight vector. 
$\sigma$ denotes sigmoid. 
$\odot$ denotes the Hadamard product. 
$\mathbf{W}_{\mathbf{h}}^{(l)}$ and $\mathbf{b}_{\mathbf{h}}^{(l)}$ are trainable parameters. 
After $L$-layer Highway GAT updating, $\mathbf{h}_{i}^{(L)}$ is the final output of this submodule, which captures the interactions with its knowledge-related $L$-hop neighbors. 

\subsubsection{Self-matching (Optional)}

Previous submodules enhance the ability of commonsense reasoning, but they do not directly address MRC. 
Therefore, we need a final step to adapt the knowledge-enhanced information to the specific MRC task. 
To achieve this, we design a self-matching submodule at the end of PIECER. 

We use a transformer block~\citep{transformer} to implement this submodule, which is composed of a multi-head self-attention layer and two fully connected layers: 
\begin{align*}
    \mathbf{o}^{\prime}_{1:N} &= \operatorname{SelfAttention} (\mathbf{h}_{1:N}^{(L)}) + \mathbf{h}_{1:N}^{(L)}, \\
    \mathbf{o}_i &= \operatorname{FC_2} ( \operatorname{ReLU} (\operatorname{FC_1} (\mathbf{o}^{\prime}_{i}) ) ) + \mathbf{o}^{\prime}_{i},
\end{align*}
where $\mathbf{h}_{1:N}^{(L)}$ are knowledge-enhanced representations of all passage tokens. 
$\operatorname{FC_x}$ denotes fully connected layers. 
$\mathbf{o}_i$ is the final output of PIECER. 

Note that PIECER is a plug-and-play module working by plugging into other MRC models. 
If an MRC model has its own matching module, we do not need another one in PIECER. 
Therefore, this submodule is optional, depending on the specific to-plug MRC model. 

\subsection{Plugging PIECER into MRC Models}

As a plug-and-play module, PIECER has high generalizability since it can be plugged into suitable positions in any MRC model. 
Figure~\ref{fig:model}~(b) demonstrates the usage of plugging in PIECER. 
For an MRC model, we first need to split it into two sequential parts: 
(1) The preceding part that takes $\mathcal{P}$ and $\mathcal{Q}$ as inputs and outputs their representations.  
(2) The subsequent part that takes these representations as inputs and predicts the final answer. 
Then, we can plug PIECER between them. 
For example, we can plug PIECER after the embedding layer or before the prediction layer for any MRC model. 

\begin{table*}[t]
\centering
\setlength{\tabcolsep}{8pt}
\begin{tabular}{l | c c c c | c c c c }
\toprule
\multirow{2}{*}{\textbf{Model}} & \multicolumn{4}{c|}{\textbf{Dev}} & \multicolumn{4}{c}{\textbf{Test}} \\ 
& \textbf{EM} & \textbf{$\Delta$ EM} & \textbf{F1} & \textbf{$\Delta$ F1} & \textbf{EM} & \textbf{$\Delta$ EM} & \textbf{F1} & \textbf{$\Delta$ F1}  \\ 
\midrule
\midrule
QANet & 36.79 & - & 37.32 & - & 38.4 & - & 38.9 & - \\
QANet + PIECER & 39.69 & 2.90$\uparrow$ & 40.20 & 2.88$\uparrow$ & 40.6 & 2.2$\uparrow$ & 41.1 & 2.2$\uparrow$ \\
\midrule
BERT$_{\text{base}}$ & 62.12 & - & 62.76 & - & 62.2 & - & 62.8 & - \\
BERT$_{\text{base}}$ + PIECER & 63.40 & 1.28$\uparrow$ & 64.01 & 1.25$\uparrow$ & 63.2 & 1.0$\uparrow$ & 63.8 & 1.0$\uparrow$ \\
\midrule
BERT$_{\text{large}}$ & 71.86 & - & 72.55 & - & 72.4 & - & 72.9 & - \\
BERT$_{\text{large}}$ + PIECER & 72.39 & 0.53$\uparrow$ & 73.04 & 0.49$\uparrow$ & 73.6 & 1.2$\uparrow$ & 74.3 & 1.4$\uparrow$ \\
\midrule
RoBERTa$_{\text{base}}$ & 78.89 & - & 79.52 & - & 79.7 & - & 80.3 & - \\
RoBERTa$_{\text{base}}$ + PIECER & 79.42 & 0.53$\uparrow$ & 80.04 & 0.52$\uparrow$ & 80.1 & 0.4$\uparrow$ & 80.7 & 0.4$\uparrow$ \\
\bottomrule
\end{tabular}
\caption{
Main evaluation results on ReCoRD. 
$\Delta$ EM and $\Delta$ F1 denote the improvements of EM and F1 introduced by PIECER, respectively. 
Results on the test set are returned by the SuperGLUE online evaluation system. 
}
\label{tab:record_main}
\end{table*}

\section{Experiments}

In this section, we describe our experimental settings and provide experimental results and analysis. 

\subsection{Datasets}

We conduct experiments on ReCoRD~\citep{record}, a large-scale public MRC dataset requiring commonsense reasoning. 
ReCoRD contains a total of 120,730 examples, 75\% of which require commonsense reasoning and is split into training, development, and test set with 100,730, 10,000, and 10,000 examples, respectively. 
Each example contains a context passage and a query, where the passage has 169.3 tokens on average and the query has 21.4 tokens on average. 
Given an example, ReCoRD requires predicting an entity span in the context passage as the answer, which can replace the entity placeholder in the query, as shown in Figure~\ref{fig:example}. 

For the external commonsense knowledge, we select a large-scale commonsense knowledge graph ConceptNet~\citep{conceptnet}. 
ConceptNet contains 34 relation categories, over 21 million relational facts, and over 8 million nodes. 
In this paper, we use an English subset with 1,165,190 nodes and 3,423,004 relational facts since ReCoRD is an English dataset. 

\subsection{Base Models}

To validate the effectiveness and generalizability of PIECER as a plug-and-play module, we plug PIECER into four representative MRC models as follows:
(1) \textbf{QANet}~\citep{qanet} is an outstanding MRC model without using PTMs. 
It contains five modules for embedding, encoding, passage-query attention, self-matching, and prediction. 
(2) BERT~\citep{bert} is one of the most widely-used PTMs. 
It uses multi-layer bidirectional transformers~\citep{transformer} as the encoder, and uses Masked Language Models (MLM) and Next Sentence Prediction (NSP) as pre-training tasks. 
We select both \textbf{BERT$_\text{base}$} and \textbf{BERT$_\text{Large}$} as our base models. 
They share the same design but are pre-trained under different configurations. 
(3) RoBERTa~\citep{roberta} is a modified PTM based on BERT. 
Firstly, it improves the MLM task and removes the NSP task. 
Secondly, it is pre-trained on larger corpora with a larger batch size for a longer time. 
Due to the limitation of our computing resources, we select only \textbf{RoBERTa$_\text{base}$} as the base model. 
For MRC, both BERT and RoBERTa are followed by a linear prediction layer to predict the answer span. 

\subsection{Experimental Settings}

For pre-trained knowledge embeddings, we select TransE~\citep{transe} implemented by OpenKE~\citep{openke} as the pre-training method, use Adam~\citep{adam} with an initial learning rate of $10^{-5}$ as the optimizer, set the knowledge embedding dimension to $100$, and pre-train for $10,000$ epochs. 

For PIECER, we tune hyper-parameters on the development set. 
For a fair comparison, we first tune the base models to achieve the best performance, and then fix their hyper-parameters before tuning PIECER. 
Generally, we use AdamW~\citep{adamw} with $\beta1=0.9$, $\beta2=0.98$, $\text{weight decay}=0.01$ as the optimizer, apply exponential moving average with a decay rate of $0.9999$ on trainable parameters, set all dropout rates to $0.1$, set the number of Highway GAT layers to $3$, and set the number of attention heads to $4$. 
Other key hyper-parameters, including learning rate, hidden dimension, and batch size, are different for each base model, and we provide details in Appendix A. 

\subsection{Evaluation Metrics}

Following previous works, we use Exact Match (EM) and F1 as the evaluation metrics. 
EM measures the percentage of the predicted answers that exactly match the ground-truth answers. 
F1 measures the token overlapping level between the predicted answers and the ground-truth answers. 
Both metrics ignore punctuations and articles. 
Since the test set is not publicly available, we use the SuperGLUE~\citep{superglue} online evaluation system\footnote{https://super.gluebenchmark.com/} to obtain our test evaluation results. 

\subsection{Main Results}

We show in Table~\ref{tab:record_main} the main evaluation results on ReCoRD. 
For each of the four base models, we compare the performance of the original model and the PIECER-plugged version (denoted as \textit{X + PIECER}). 
From the table, we have the following observations: 
(1) PIECER introduces stable EM and F1 improvements for all base models. 
This validates the effectiveness of PIECER to enhance MRC that requires commonsense reasoning. 
(2) Comparing the performance improvements on four base models, QANet benefits the most from PIECER as an MRC model without using PTMs, while PTMs like BERT and RoBERTa gain moderate improvements. 
We speculate that this difference may be due to the knowledge overlaps between PTMs and PIECER: PTMs have already encoded certain knowledge in their representations implicitly, which may overlap with commonsense knowledge introduced by PIECER. 
(3) 
We further investigate three PTMs that may contain overlapping knowledge, and observe that PIECER still introduces stable performance improvements. 
This is due to the difference in how to leverage knowledge: PTMs encode knowledge in an uncontrollable and implicit way, while PIECER can actively select related commonsense knowledge stored in a KG for explicit use. 

\subsection{Analysis and Discussions}

In this subsection, we provide elaborate analyses of PIECER to further validate its effectiveness and explore its properties. 
Since the test set is not publicly available and the number of submissions to the online evaluation system is restricted, all analysis experiments are based on the development set. 

\begin{table}[t]
\centering
\setlength{\tabcolsep}{1.5pt}
\begin{tabular}{l | c c}
\toprule
\textbf{Model} & \textbf{EM} & \textbf{F1} \\
\midrule
\midrule
RoBERTa$_{\text{base}}$ + PIECER & 79.42 & 80.04 \\
w/o self-matching & 79.41 & 79.99 \\
w/o knowledge embedding injection & 79.25 & 79.92 \\
w/o knowledge-guided reasoning & 78.98 & 79.67 \\
RoBERTa$_\text{base}$ & 78.89 & 79.52 \\
\bottomrule
\end{tabular}
\caption{
Ablation study of PIECER on the development set of ReCoRD based on RoBERTa$_\text{base}$. 
}
\label{tab:ablation}
\end{table}

\subsubsection{Ablation Study}

To validate the effectiveness of each submodule in PIECER, we provide ablation experimental results of PIECER on the development set of ReCoRD in Table~\ref{tab:ablation}. 
We select RoBERTa$_\text{base}$ + PIECER as the baseline and attempt to remove the self-matching submodule, the knowledge embedding injection submodule, the knowledge-guided reasoning submodule, and the whole PIECER, respectively. 
Comparing the results of RoBERTa$_\text{base}$ + PIECER with three simplified versions, we observe that removing any submodule will result in a performance drop, especially for the knowledge-guided reasoning submodule. 
This suggests that each submodule is necessary for PIECER, and the connection information in a KG is more essential than knowledge embeddings for commonsense reasoning. 
Further, comparing the results of RoBERTa$_\text{base}$ with three simplified versions of PIECER, we observe that the performance of each simplified version is higher than that without the whole PIECER. 
This indicates that the base model can benefit from each submodule. 

\begin{figure}[t]
\centering
\includegraphics[width=0.99\linewidth]{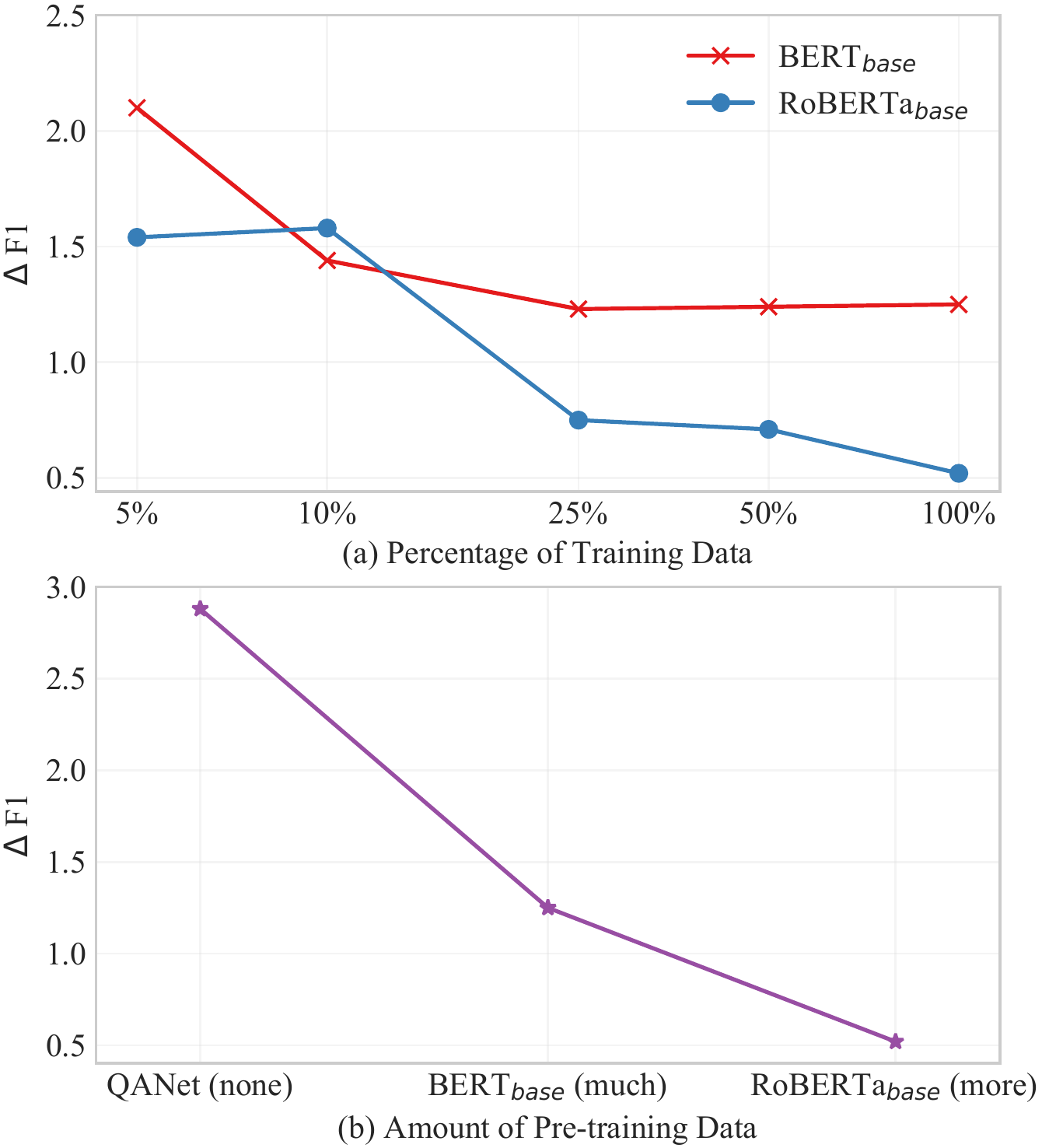}
\caption{
F1 improvements introduced by PIECER in different training  (a) / pre-training (b) resource settings. 
}
\label{fig:low-resource}
\end{figure}

\subsubsection{PIECER in Low-resource Settings}

Since PIECER has the ability to leverage external commonsense knowledge, we hypothesize that it can alleviate the problem of data insufficiency. 
To verify our hypothesis, we compare the performance of PIECER in different resource settings. 
Firstly, we study how the amount of training data influences its effectiveness, using the F1 improvement introduced by PIECER (denoted by $\Delta$F1) as the metric. 
Figure~\ref{fig:low-resource}~(a) shows the results: 
for both BERT$_{\text{base}}$ and RoBERTa$_{\text{base}}$, PIECER can introduce more improvements when the amount of training data is smaller, i.e., with lower training resources. 
Secondly, we study how the amount of pre-training data influences the effectiveness of PIECER and show the results in Figure~\ref{fig:low-resource}~(b). 
PIECER performs the best based on QANet that does not use any pre-training data, and the worst based on RoBERTa$_{\text{base}}$ that uses the most pre-training data. 
This reveals that PIECER can also introduce more improvements with lower pre-training resources. 

As indicated by the above results, PIECER is especially effective with low training or pre-training resources. 
Thus, plugging in PIECER could be the simplest and cheapest solution when facing insufficient training data or the incapability to conduct large-scale pre-training. 

\begin{table}[t]
\centering
\setlength{\tabcolsep}{1.5pt}
\begin{tabular}{l | c c}
\toprule
\textbf{Method (Pre-training Epochs)} & \textbf{EM} & \textbf{F1} \\
\midrule
\midrule
DistMult (1,000 epochs) & 38.82 & 39.27 \\
DistMult (5,000 epochs) & 39.09 & 39.55 \\
TransE (1,000 epochs) & 39.64 & 40.15 \\
TransE (10,000 epochs) & 39.69 & 40.20 \\
\midrule
w/o knowledge embedding injection & 39.18 & 39.66 \\
\bottomrule
\end{tabular}
\caption{Performance with different knowledge embeddings pre-trained under different configurations. }
\label{tab:kge}
\end{table}

\subsubsection{Robustness to Different Knowledge Embeddings}

To get a deeper understanding of the impact of knowledge embeddings on PIECER, we compare different pre-training ways. 
Table~\ref{tab:kge} shows the experimental results based on QANet+PIECER. 
From the table, we observe that: 
(1) A proper KGE method is essential, since the performance with knowledge embeddings pre-trained by DistMult~\citep{distmult} is even worse than that without using knowledge embeddings (\textit{w/o knowledge embedding injection}). 
Even so, the impact of different KGE methods is not significant for PIECER. 
(2) PIECER is robust to the hyper-parameters, such as pre-training epochs, since for both DistMult and TransE, the number of epochs has little influence on the performance. 

The above observations reveal the robustness of PIECER to KGE methods and hyper-parameters: 
PIECER can keep a high performance even with a sub-optimal configuration, since PIECER mainly benefits from the connection information instead of knowledge embeddings.  
By contrast, methods that leverage only knowledge embeddings are sensitive to KGE configurations and have to spend extensive efforts to search for the optimal. 
The robustness of PIECER to the pre-training KGE configuration can ease the burden of hyper-parameter tuning.

\begin{table}[t]
\centering
\setlength{\tabcolsep}{15pt}
\begin{tabular}{l | c c}
\toprule
\textbf{Model} & \textbf{EM} & \textbf{F1} \\
\midrule
\midrule
Highway GAT & 63.40 & 64.01 \\
Res GAT & 62.54 & 63.16 \\
w/o highway & 58.39 & 59.46 \\
\midrule
Highway GCN & 62.64 & 63.33 \\
Highway GIN & 62.66 & 63.29 \\
\bottomrule
\end{tabular}
\caption{Performance of different GCN architectures. }
\label{tab:gcn}
\end{table}

\begin{figure}[t]
\centering
\includegraphics[width=0.99\linewidth]{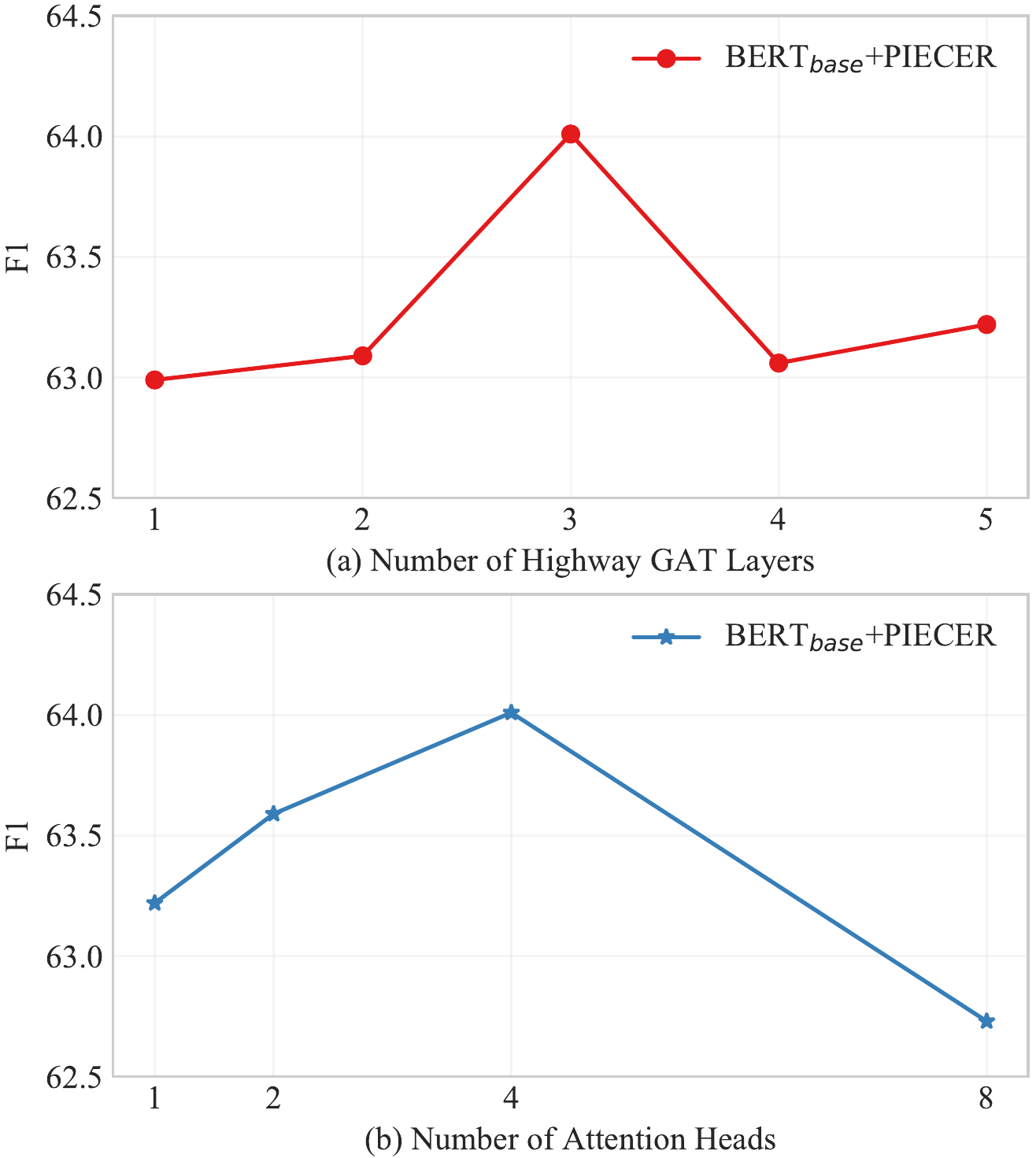}
\caption{
Performance with different numbers of Highway GAT layers (a) / attention heads (b). 
}
\label{fig:layer-head}
\end{figure}

\subsubsection{Impact of GCN Architectures}

To validate the design of the Highway GAT in PIECER, we compare other GCN architectures with it based on BERT$_{\text{base}}$ + PIECER. 
Firstly, we study the effectiveness of the highway connection. 
As shown in Table~\ref{tab:gcn}, if we replace the highway connection by a simple residual connection (denoted by \textit{Res GAT}), the performance will drop slightly. 
Further, if we remove the whole highway connection (denoted by \textit{w/o highway}), the performance will be severely degraded. 
This suggests that the highway connection plays a key role in our Highway GAT. 
Secondly, we study the influence of basic GCN models. 
We replace GAT in our module by GCN~\citep{gcn} or GIN~\citep{gin}, and keep the highway connection for both of them for a fair comparison. 
Table~\ref{tab:gcn} shows that both Highway GCN and Highway GIN perform worse than Highway GAT. 
Thirdly, we study how the number of Highway GAT layers and attention heads influences the performance. 
As shown in Figure~\ref{fig:layer-head}~(a), Highway GAT achieves the peak performance with 3 layers. 
We speculate that too few layers may build too short reasoning chains, while too many layers may lead to the over-smoothing problem~\citep{over_smoothing}. 
Figure~\ref{fig:layer-head}~(b) shows that Highway GAT achieves the best performance with 4 heads, which reveals that the attention heads are not the more the better. 

\begin{table}[t]
\centering
\setlength{\tabcolsep}{8pt}
\begin{tabular}{l | c c}
\toprule
\textbf{Graph} & \textbf{EM} & \textbf{F1} \\
\midrule
\midrule
joint query-passage graph & 63.40 & 64.01 \\
w/o coreference edge & 63.00 & 63.65 \\
w/o knowledge edge & 62.34 & 62.98 \\
\midrule
w/ complete graph & 62.33 & 62.94 \\
\bottomrule
\end{tabular}
\caption{Performance with joint query-passage graphs constructed in different ways. }
\label{tab:graph}
\end{table}

\subsubsection{Impact of Graph Construction Methods}

To study the impact of the joint query-passage graph, we compare different ways to construct the joint graph. 
In Table~\ref{tab:graph}, we provide the results based on BERT$_{\text{base}}$ + PIECER. 
From the table, we have the following observations: 
(1) Removing either the coreference edge or the knowledge edge degrades the performance. 
This verifies the necessity of these two categories of edges. 
(2) A complete graph performs worse than the joint query-passage graph in PIECER, and even worse than that with only the coreference or knowledge edge. 
This suggests that the effectiveness of the knowledge-guided reasoning submodule in PIECER is due to not only the Highway GAT reasoning model, but also the knowledge-oriented connections that can explicitly guide the reasoning chain building. 

\section{Related Work}

MRC is a longstanding task in NLP.
In recent years, deep learning based methods such as Match-LSTM~\citep{match_lstm}, BiDAF~\citep{bidaf}, DCN~\citep{dcn}, R-Net~\citep{rnet}, and QANet~\citep{qanet} have become the mainstream. 
They share a similar paradigm, composed of an embedding module, an encoding module, an attention-based interaction module, a self-matching module, and a prediction module. 
As large-scale PTMs such as ELMo~\citep{elmo}, GPT~\citep{gpt}, BERT~\citep{bert}, and RoBERTa~\citep{roberta} are proposed, MRC achieves further advance by adopting PTMs. 
However, their ability of commonsense reasoning remains a question. 

As MRC datasets requiring knowledge such as ReCoRD are proposed, various methods attempt to introduce external knowledge stored in KGs such as ConceptNet~\citep{conceptnet}, NELL~\citep{nell}, and WordNet~\citep{wordnet} into MRC. 
\citet{knreader} encode knowledge as a key-value memory and enrich each word by memory querying. 
\citet{kblstm} and~\citet{ktnet} enrich each word by applying attention mechanism to its KG neighbors and a sentinel vector. 
\citet{skg} update the representation of each word by aggregating knowledge embeddings of its KG neighbors. 
These methods enrich each word separately, ignoring the knowledge-oriented connections between words, while these connections form a main part to compose a KG and could be pivotal cues to build the commonsense reasoning chains. 

Graph Convolutional Networks (GCN) are effective to deal with graph-like data. 
\citet{gcn} propose a fast approximate convolution method, which becomes one of the most popular spectral GCN models. 
Besides spectral GCNs, spatial GCNs such as GraphSAGE~\citep{graphsage}, GIN~\citep{gin}, and GAT~\citep{gat} form another category of GCNs. 
In recent years, GCNs have been applied to various NLP tasks such as Relation Extraction~\citep{gcn_re1,gcn_re2}, Natural Language Inference~\citep{gcn_nli}, and Machine Reading Comprehension~\citep{gcn_mrc1,gcn_mrc2,skg}. 
However, existing GCN-based MRC methods do not thoroughly investigate the knowledge-oriented connections between words or explicitly incorporate them. 

\section{Conclusion}

In this paper, we propose to enhance commonsense reasoning in MRC by explicitly incorporating the connection information in KGs. 
For high generalizability, we design a plug-and-play module called PIECER, which can be plugged into suitable positions in any MRC model. 
Experimental results on ReCoRD validate the effectiveness of PIECER by plugging it into four representative base MRC models. 
Further analysis reveals that PIECER is especially effective in low-resource settings. 

\bibliographystyle{acl_natbib}
\bibliography{anthology,acl2021}

\begin{thebibliography}{41}
\expandafter\ifx\csname natexlab\endcsname\relax\def\natexlab#1{#1}\fi

\bibitem[{Bordes et~al.(2013)Bordes, Usunier, Garc{\'{\i}}a{-}Dur{\'{a}}n,
  Weston, and Yakhnenko}]{transe}
Antoine Bordes, Nicolas Usunier, Alberto Garc{\'{\i}}a{-}Dur{\'{a}}n, Jason
  Weston, and Oksana Yakhnenko. 2013.
\newblock Translating embeddings for modeling multi-relational data.
\newblock In \emph{{NeurIPS} 2013}, pages 2787--2795.

\bibitem[{Cao et~al.(2019)Cao, Aziz, and Titov}]{gcn_mrc1}
Nicola~De Cao, Wilker Aziz, and Ivan Titov. 2019.
\newblock Question answering by reasoning across documents with graph
  convolutional networks.
\newblock In \emph{{NAACL-HLT} 2019}, pages 2306--2317.

\bibitem[{Carlson et~al.(2010)Carlson, Betteridge, Kisiel, Settles, Jr., and
  Mitchell}]{nell}
Andrew Carlson, Justin Betteridge, Bryan Kisiel, Burr Settles, Estevam
  R.~Hruschka Jr., and Tom~M. Mitchell. 2010.
\newblock Toward an architecture for never-ending language learning.
\newblock In \emph{{AAAI} 2010}, pages 1306--1313.

\bibitem[{Dai et~al.(2020)Dai, Ren, Zeng, Chang, and Sui}]{gcn_re2}
Damai Dai, Jing Ren, Shuang Zeng, Baobao Chang, and Zhifang Sui. 2020.
\newblock Coarse-to-fine entity representations for document-level relation
  extraction.
\newblock \emph{CoRR}, abs/2012.02507.

\bibitem[{Devlin et~al.(2019)Devlin, Chang, Lee, and Toutanova}]{bert}
Jacob Devlin, Ming{-}Wei Chang, Kenton Lee, and Kristina Toutanova. 2019.
\newblock {BERT:} pre-training of deep bidirectional transformers for language
  understanding.
\newblock In \emph{{NAACL-HLT} 2019}, pages 4171--4186.

\bibitem[{Guo et~al.(2019)Guo, Zhang, and Lu}]{gcn_re1}
Zhijiang Guo, Yan Zhang, and Wei Lu. 2019.
\newblock Attention guided graph convolutional networks for relation
  extraction.
\newblock In \emph{{ACL} 2019}, pages 241--251.

\bibitem[{Hamilton et~al.(2017)Hamilton, Ying, and Leskovec}]{graphsage}
William~L. Hamilton, Zhitao Ying, and Jure Leskovec. 2017.
\newblock Inductive representation learning on large graphs.
\newblock In \emph{{NeurIPS} 2017}, pages 1024--1034.

\bibitem[{Han et~al.(2018)Han, Cao, Lv, Lin, Liu, Sun, and Li}]{openke}
Xu~Han, Shulin Cao, Xin Lv, Yankai Lin, Zhiyuan Liu, Maosong Sun, and Juanzi
  Li. 2018.
\newblock Openke: An open toolkit for knowledge embedding.
\newblock In \emph{{EMNLP} 2018: System Demonstrations}, pages 139--144.

\bibitem[{Jia and Liang(2017)}]{pattern_matching1}
Robin Jia and Percy Liang. 2017.
\newblock Adversarial examples for evaluating reading comprehension systems.
\newblock In \emph{{EMNLP} 2017}, pages 2021--2031.

\bibitem[{Kaushik and Lipton(2018)}]{pattern_matching3}
Divyansh Kaushik and Zachary~C. Lipton. 2018.
\newblock How much reading does reading comprehension require? {A} critical
  investigation of popular benchmarks.
\newblock In \emph{{EMNLP} 2018}, pages 5010--5015.

\bibitem[{Kingma and Ba(2015)}]{adam}
Diederik~P. Kingma and Jimmy Ba. 2015.
\newblock Adam: {A} method for stochastic optimization.
\newblock In \emph{{ICLR} 2015}.

\bibitem[{Kipf and Welling(2017)}]{gcn}
Thomas~N. Kipf and Max Welling. 2017.
\newblock Semi-supervised classification with graph convolutional networks.
\newblock In \emph{{ICLR} 2017}.

\bibitem[{Li et~al.(2018)Li, Han, and Wu}]{over_smoothing}
Qimai Li, Zhichao Han, and Xiao{-}Ming Wu. 2018.
\newblock Deeper insights into graph convolutional networks for semi-supervised
  learning.
\newblock In \emph{{AAAI} 2018}, pages 3538--3545.

\bibitem[{Liu et~al.(2019)Liu, Ott, Goyal, Du, Joshi, Chen, Levy, Lewis,
  Zettlemoyer, and Stoyanov}]{roberta}
Yinhan Liu, Myle Ott, Naman Goyal, Jingfei Du, Mandar Joshi, Danqi Chen, Omer
  Levy, Mike Lewis, Luke Zettlemoyer, and Veselin Stoyanov. 2019.
\newblock Roberta: {A} robustly optimized {BERT} pretraining approach.
\newblock \emph{CoRR}, abs/1907.11692.

\bibitem[{Loshchilov and Hutter(2019)}]{adamw}
Ilya Loshchilov and Frank Hutter. 2019.
\newblock Decoupled weight decay regularization.
\newblock In \emph{{ICLR} 2019}.

\bibitem[{Mihaylov and Frank(2018)}]{knreader}
Todor Mihaylov and Anette Frank. 2018.
\newblock Knowledgeable reader: Enhancing cloze-style reading comprehension
  with external commonsense knowledge.
\newblock In \emph{{ACL} 2018}, pages 821--832.

\bibitem[{Miller(1995)}]{wordnet}
George~A. Miller. 1995.
\newblock Wordnet: {A} lexical database for english.
\newblock \emph{Commun. {ACM}}, 38(11):39--41.

\bibitem[{Peters et~al.(2018)Peters, Neumann, Iyyer, Gardner, Clark, Lee, and
  Zettlemoyer}]{elmo}
Matthew~E. Peters, Mark Neumann, Mohit Iyyer, Matt Gardner, Christopher Clark,
  Kenton Lee, and Luke Zettlemoyer. 2018.
\newblock Deep contextualized word representations.
\newblock In \emph{{NAACL-HLT} 2018}, pages 2227--2237.

\bibitem[{Qiu et~al.(2019)Qiu, Zhang, Feng, Liao, Jiang, Lyu, Liu, and
  Zhao}]{skg}
Delai Qiu, Yuanzhe Zhang, Xinwei Feng, Xiangwen Liao, Wenbin Jiang, Yajuan Lyu,
  Kang Liu, and Jun Zhao. 2019.
\newblock Machine reading comprehension using structural knowledge graph-aware
  network.
\newblock In \emph{{EMNLP-IJCNLP} 2019}, pages 5895--5900.

\bibitem[{Radford et~al.(2018)Radford, Narasimhan, Salimans, and
  Sutskever}]{gpt}
Alec Radford, Karthik Narasimhan, Tim Salimans, and Ilya Sutskever. 2018.
\newblock Improving language understanding by generative pre-training.
\newblock \emph{OpenAI Technical Report}.

\bibitem[{Rajpurkar et~al.(2018)Rajpurkar, Jia, and Liang}]{pattern_matching2}
Pranav Rajpurkar, Robin Jia, and Percy Liang. 2018.
\newblock Know what you don't know: Unanswerable questions for squad.
\newblock In \emph{{ACL} 2018}, pages 784--789.

\bibitem[{Rajpurkar et~al.(2016)Rajpurkar, Zhang, Lopyrev, and Liang}]{squad}
Pranav Rajpurkar, Jian Zhang, Konstantin Lopyrev, and Percy Liang. 2016.
\newblock Squad: 100, 000+ questions for machine comprehension of text.
\newblock In \emph{{EMNLP} 2016}, pages 2383--2392.

\bibitem[{Seo et~al.(2017)Seo, Kembhavi, Farhadi, and Hajishirzi}]{bidaf}
Min~Joon Seo, Aniruddha Kembhavi, Ali Farhadi, and Hannaneh Hajishirzi. 2017.
\newblock Bidirectional attention flow for machine comprehension.
\newblock In \emph{{ICLR} 2017}.

\bibitem[{Song et~al.(2018)Song, Wang, Yu, Zhang, Florian, and
  Gildea}]{gcn_mrc2}
Linfeng Song, Zhiguo Wang, Mo~Yu, Yue Zhang, Radu Florian, and Daniel Gildea.
  2018.
\newblock Exploring graph-structured passage representation for multi-hop
  reading comprehension with graph neural networks.
\newblock \emph{CoRR}, abs/1809.02040.

\bibitem[{Speer et~al.(2017)Speer, Chin, and Havasi}]{conceptnet}
Robyn Speer, Joshua Chin, and Catherine Havasi. 2017.
\newblock Conceptnet 5.5: An open multilingual graph of general knowledge.
\newblock In \emph{{AAAI} 2017}, pages 4444--4451.

\bibitem[{Srivastava et~al.(2015)Srivastava, Greff, and Schmidhuber}]{highway}
Rupesh~Kumar Srivastava, Klaus Greff, and J{\"{u}}rgen Schmidhuber. 2015.
\newblock Highway networks.
\newblock \emph{CoRR}, abs/1505.00387.

\bibitem[{Trischler et~al.(2017)Trischler, Wang, Yuan, Harris, Sordoni,
  Bachman, and Suleman}]{newsqa}
Adam Trischler, Tong Wang, Xingdi Yuan, Justin Harris, Alessandro Sordoni,
  Philip Bachman, and Kaheer Suleman. 2017.
\newblock Newsqa: {A} machine comprehension dataset.
\newblock In \emph{Rep4NLP@ACL 2017}, pages 191--200.

\bibitem[{Vaswani et~al.(2017)Vaswani, Shazeer, Parmar, Uszkoreit, Jones,
  Gomez, Kaiser, and Polosukhin}]{transformer}
Ashish Vaswani, Noam Shazeer, Niki Parmar, Jakob Uszkoreit, Llion Jones,
  Aidan~N. Gomez, Lukasz Kaiser, and Illia Polosukhin. 2017.
\newblock Attention is all you need.
\newblock In \emph{{NeurIPS} 2017}, pages 5998--6008.

\bibitem[{Velickovic et~al.(2018)Velickovic, Cucurull, Casanova, Romero,
  Li{\`{o}}, and Bengio}]{gat}
Petar Velickovic, Guillem Cucurull, Arantxa Casanova, Adriana Romero, Pietro
  Li{\`{o}}, and Yoshua Bengio. 2018.
\newblock Graph attention networks.
\newblock In \emph{{ICLR} 2018}.

\bibitem[{Wang et~al.(2019)Wang, Pruksachatkun, Nangia, Singh, Michael, Hill,
  Levy, and Bowman}]{superglue}
Alex Wang, Yada Pruksachatkun, Nikita Nangia, Amanpreet Singh, Julian Michael,
  Felix Hill, Omer Levy, and Samuel~R. Bowman. 2019.
\newblock Superglue: {A} stickier benchmark for general-purpose language
  understanding systems.
\newblock In \emph{{NeurIPS} 2019,}, pages 3261--3275.

\bibitem[{Wang and Jiang(2017)}]{match_lstm}
Shuohang Wang and Jing Jiang. 2017.
\newblock Machine comprehension using match-lstm and answer pointer.
\newblock In \emph{{ICLR} 2017}.

\bibitem[{Wang et~al.(2017)Wang, Yang, Wei, Chang, and Zhou}]{rnet}
Wenhui Wang, Nan Yang, Furu Wei, Baobao Chang, and Ming Zhou. 2017.
\newblock Gated self-matching networks for reading comprehension and question
  answering.
\newblock In \emph{{ACL} 2017}, pages 189--198.

\bibitem[{Wang et~al.(2020)Wang, Li, and Zeng}]{gcn_nli}
Zikang Wang, Linjing Li, and Daniel Zeng. 2020.
\newblock Knowledge-enhanced natural language inference based on knowledge
  graphs.
\newblock In \emph{{COLING} 2020}, pages 6498--6508.

\bibitem[{Xiong et~al.(2017)Xiong, Zhong, and Socher}]{dcn}
Caiming Xiong, Victor Zhong, and Richard Socher. 2017.
\newblock Dynamic coattention networks for question answering.
\newblock In \emph{{ICLR} 2017}.

\bibitem[{Xu et~al.(2019)Xu, Hu, Leskovec, and Jegelka}]{gin}
Keyulu Xu, Weihua Hu, Jure Leskovec, and Stefanie Jegelka. 2019.
\newblock How powerful are graph neural networks?
\newblock In \emph{{ICLR} 2019}.

\bibitem[{Yang et~al.(2019)Yang, Wang, Liu, Liu, Lyu, Wu, She, and Li}]{ktnet}
An~Yang, Quan Wang, Jing Liu, Kai Liu, Yajuan Lyu, Hua Wu, Qiaoqiao She, and
  Sujian Li. 2019.
\newblock Enhancing pre-trained language representations with rich knowledge
  for machine reading comprehension.
\newblock In \emph{{ACL} 2019}, pages 2346--2357.

\bibitem[{Yang and Mitchell(2017)}]{kblstm}
Bishan Yang and Tom~M. Mitchell. 2017.
\newblock Leveraging knowledge bases in lstms for improving machine reading.
\newblock In \emph{{ACL} 2017}, pages 1436--1446.

\bibitem[{Yang et~al.(2015)Yang, Yih, He, Gao, and Deng}]{distmult}
Bishan Yang, Wen{-}tau Yih, Xiaodong He, Jianfeng Gao, and Li~Deng. 2015.
\newblock Embedding entities and relations for learning and inference in
  knowledge bases.
\newblock In \emph{{ICLR} 2015}.

\bibitem[{Yu et~al.(2018)Yu, Dohan, Luong, Zhao, Chen, Norouzi, and Le}]{qanet}
Adams~Wei Yu, David Dohan, Minh{-}Thang Luong, Rui Zhao, Kai Chen, Mohammad
  Norouzi, and Quoc~V. Le. 2018.
\newblock Qanet: Combining local convolution with global self-attention for
  reading comprehension.
\newblock In \emph{{ICLR} 2018}.

\bibitem[{Zhang et~al.(2018)Zhang, Liu, Liu, Gao, Duh, and Durme}]{record}
Sheng Zhang, Xiaodong Liu, Jingjing Liu, Jianfeng Gao, Kevin Duh, and
  Benjamin~Van Durme. 2018.
\newblock Record: Bridging the gap between human and machine commonsense
  reading comprehension.
\newblock \emph{CoRR}, abs/1810.12885.

\bibitem[{Zhou et~al.(2018)Zhou, Young, Huang, Zhao, Xu, and
  Zhu}]{graph_attention}
Hao Zhou, Tom Young, Minlie Huang, Haizhou Zhao, Jingfang Xu, and Xiaoyan Zhu.
  2018.
\newblock Commonsense knowledge aware conversation generation with graph
  attention.
\newblock In \emph{{IJCAI} 2018}, pages 4623--4629.

\end{thebibliography}

\appendix
\appendixpage

\section{Details of Experimental Settings}

We tune hyper-parameters on the development set. 
For each base model, we first tune its hyper-parameters to achieve the best performance and then fix its best hyper-parameters before tuning PIECER. 
The criterion for selecting the best hyper-parameters is the F1 on the development set. 
Details of the general hyper-parameters and hyper-parameters for each base model are described as follows. 

\textbf{General hyper-parameters: }
(1) Empirically, we use AdamW with $\beta1=0.9$, $\beta2=0.98$, $\text{eps}=10^{-6}$, $\text{weight decay}=0.01$ as the optimizer. 
(2) Empirically, we adopt a slanted triangular learning rate scheduler, which first linearly increases the learning rate from $0$ to the peak value during the first $6\%$ steps, and then linearly decreases it to $0$ during the remaining steps.  
(3) Empirically, we apply exponential moving average with a decay rate of $0.9999$ on trainable parameters. 
(4) Empirically, we set all dropout rates to $0.1$. 
(5) We try the number of Highway GAT layers to in $\{1, 2, 3, 4, 5\}$, and finally select $3$. 
(6) We try the number of attention heads in $\{1, 2, 4, 8\}$, and finally select $4$. 

\textbf{Hyper-parameters for QANet: }
(1) We try the peak learning rate in $\{0.0005, 0.001, 0.005\}$, and finally select $0.001$. 
(2) We try the hidden dimension in $\{64, 128, 256\}$, and finally select $128$. 
(3) We try the batch size in $\{8, 16, 32, 64\}$, and finally select $32$. 
(4) We try to plug PIECER after the embedding layer, after the encoding layer, and at both these two positions, and finally select plugging at both two positions. 
(5) Since QANet has its own self-matching layer, we remove the optional self-matching submodule in PIECER. 
(6) We train for $30$ epochs and evaluate the model on the development set after each epoch. 
Finally, we report the best F1 achieved during $30$ epochs and use the corresponding model to predict answers on the test set. 

\textbf{Hyper-parameters for BERT$_\text{base}$: }
(1) We try the peak learning rate for BERT module in $\{0.00001, 0.00003, 0.00005\}$, and finally select $0.00001$. 
(2) We try the peak learning rate for other modules in $\{0.0005, 0.001, 0.005\}$, and finally select $0.0005$. 
(3) We set the hidden dimension to $768$, the same as BERT$_\text{base}$. 
(4) We try the batch size in $\{4, 8, 16, 32\}$, and finally select $4$. 
(5) We plug PIECER between BERT$_\text{base}$ and the predicting layer since BERT$_\text{base}$ is impartible. 
(6) We keep the optional self-matching submodule in PIECER. 
(7) We train for $4$ epochs and evaluate the model on the development set after each epoch. 
Finally, we report the best F1 achieved during $4$ epochs and use the corresponding model to predict answers on the test set. 

\textbf{Hyper-parameters for BERT$_\text{large}$: }
(1) We try the peak learning rate for BERT module in $\{0.00001, 0.00003, 0.00005\}$, and finally select $0.00003$. 
(2) We try the peak learning rate for other modules in $\{0.0005, 0.001, 0.005\}$, and finally select $0.001$. 
(3) We set the hidden dimension to $1024$, the same as BERT$_\text{large}$. 
(4) We try the batch size in $\{4, 8, 16, 32\}$, and finally select $32$. 
(5) We plug PIECER between BERT$_\text{large}$ and the predicting layer since BERT$_\text{large}$ is impartible. 
(6) We keep the optional self-matching submodule in PIECER. 
(7) We train for $4$ epochs and evaluate the model on the development set after each epoch. 
Finally, we report the best F1 achieved during $4$ epochs and use the corresponding model to predict answers on the test set. 

\textbf{Hyper-parameters for RoBERTa$_\text{base}$: }
(1) We try the peak learning rate for RoBERTa module in $\{0.00001, 0.00003, 0.00005\}$, and finally select $0.00001$. 
(2) We try the peak learning rate for other modules in $\{0.0005, 0.001, 0.005\}$, and finally select $0.0005$. 
(3) We set the hidden dimension to $768$, the same as RoBERTa$_\text{base}$. 
(4) We try the batch size in $\{4, 8, 16, 32\}$, and finally select $4$. 
(5) We plug PIECER between RoBERTa$_\text{base}$ and the predicting layer since RoBERTa$_\text{base}$ is impartible. 
(6) We keep the optional self-matching submodule in PIECER. 
(7) We train for $4$ epochs and evaluate the model on the development set after each epoch. 
Finally, we report the best F1 achieved during $4$ epochs and use the corresponding model to predict answers on the test set. 

\end{document}